\documentclass[10pt,letterpaper]{article}

\usepackage{cogsci}
\usepackage{pslatex}
\usepackage{apacite}
\usepackage{rotating}
\usepackage{array}
\usepackage{multirow}
\usepackage{graphicx}
\usepackage[small]{caption}
\usepackage{subcaption}
\usepackage{xcolor, soul}
\usepackage{url}
\usepackage{booktabs}
\DeclareRobustCommand{\mycite}{\cite{sellam2022multiberts,yedetore-etal-2023-poor}}

\title{A systematic investigation of learnability from single child linguistic input}

\author{\large \bf Yulu Qin$^1$ (yq810@nyu.edu)\\ \large \bf Wentao Wang$^1$ (ww2135@nyu.edu)\\
\large \bf Brenden M. Lake$^{1,2}$ (brenden@nyu.edu)\\
\large $^1$Center for Data Science, $^2$Department of Psychology, New York University
}

\usepackage{lipsum} 
\usepackage{booktabs} 

\begin{document}

\maketitle


\begin{abstract}
Language models (LMs) have demonstrated remarkable proficiency in generating linguistically coherent text, sparking discussions about their relevance to understanding human language learnability. However, a significant gap exists between the training data for these models and the linguistic input a child receives. LMs are typically trained on data that is orders of magnitude larger and fundamentally different from child-directed speech \cite{warstadt2022artificial, warstadt2023findings, frank2023bridging}. Addressing this discrepancy, our research focuses on training LMs on subsets of a single child's linguistic input. Previously, \citeA{wang2023finding} found that LMs trained in this setting can form syntactic and semantic word clusters and develop sensitivity to certain linguistic phenomena, but they only considered LSTMs and simpler neural networks trained from just one single-child dataset. 
Here, to examine the robustness of learnability from single-child input, we systematically train six different model architectures on five datasets (3 single-child and 2 baselines). We find that the models trained on single-child datasets showed consistent results that matched with previous work, underscoring the robustness of forming meaningful syntactic and semantic representations from a subset of a child's linguistic input.

\end{abstract}

\section{Introduction}
Young children are remarkably efficient language learners, yet the mechanisms behind language acquisition remain a scientific puzzle. Meanwhile, important advances in language models (LMs) for natural language processing provide us with new, powerful computational tools to investigate fundamental questions regarding language acquisition and its relationship with human cognition \cite{warstadt2022artificial,frank2023llm}. 
Trained on trillions of written words, contemporary Transformer-based Large Language Models (LLMs) can produce coherent text with a proficiency that far exceeds the predictions of experts in the field from a decade ago \cite{chang2023language}, raising important questions about the degree to which strong inductive biases and language-specific mechanisms are needed to acquire language beyond more general distributional learning mechanisms \cite{landauer1998introduction, elman1990finding}. To improve the relevance of language models as cognitive models of human language acquisition, previous efforts trained models on aggregated linguistic input across multiple children \cite{warstadt2023findings, huebner2021babyberta}. As in several works \cite{wang2023finding, vong2024grounded, abend2017bootstrapping, waterfall2010empirical}, we train models on subsets of the linguistic input that just a single child was exposed to. Children must learn language from only their own input---they cannot share and aggregate input with others---and thus this is the setting we focus on here.

Here, we use a recent article by \citeA{wang2023finding} as a launchpad for our new learnability studies based on a single child's input. \citeA{wang2023finding} applied two neural language models, Continuous Bag-of-Words \cite<CBOW;>{mikolov2013efficient} and Long Short-Term Memory \cite<LSTM;>{hochreiter1997long}, to the SAYCam-S dataset, a longitudinal collection of transcribed linguistic inputs to a single child aged 6 to 25 months \cite{sullivan2021saycam}. \citeA{wang2023finding}'s study revealed that these models successfully recovered lexical classes that reflect key syntactic and semantic distinctions, including nouns, verbs, animals, body parts, etc., from the process of learning to predict the next word in transcribed child-directed utterances. Additionally, they employed the Zorro test suite to evaluate the models' grammatical knowledge through acceptability judgments \cite{huebner2021babyberta}. However, these promising findings are based on two model architectures trained on only one single child's data, thus limiting the generalizability of their results. Our research builds upon this groundwork by investigating the robustness of \citeA{wang2023finding}'s learnability results from one child's input across different settings, including multiple datasets and different model architectures, to see which combinations of datasets and architectures can produce successful learners.

Specifically, in this study, we examined 6 model architectures (3 model classes and 2 sizes each) trained on 5 datasets: 3 datasets representing input to individual children and 2 others representing meaningful baselines for comparison. Each combination of architecture and dataset was analyzed through linguistic acceptability tests, visualizations of word embeddings, and cloze tests. Across each of these settings, we find that the results are robust and similar to \citeA{wang2023finding}'s.

\section{Methods}
\subsection{Datasets}

\begin{table*}[!ht]
\centering
\caption{\textbf{Dataset Statistics.} SAYCam-S, Sarah, and Ellie are three single-child datasets. Note that all datasets except CHILDES have a similar number of training tokens.}
\label{tab:dataset}
\begin{tabular}{l|r|ccc|cc}

\hline
    &  & \textbf{SAYCam-S} & \textbf{Sarah} & \textbf{Ellie} & Wikipedia & CHILDES \\
\hline

\multirow{5.5}{*}{\begin{sideways}Training\end{sideways}}
& Number of utterances         & 26,322 &  32,965 & 38,140 & 10,504 & 1,151,816 \\
& Mean (SD) utterance length    & 8.06 (5.46) &   6.71 (3.32) & 6.29 (3.14) & 24.81 (14.60) & 7.09 (4.19) \\
& Number of tokens             & 212,064 &   221,211 & 239,807 & 260,580 & 8,163,820 \\
& Out-of-vocabulary rate       & 1.85\% &   1.26\% & 1.74\% & 9.69\%  & 0.26\%\\
& Vocabulary size              & 2350 &   2333 & 2780 & 8833 & 15,762\\
\hline
\multirow{4.4}{*}{\begin{sideways}Validation\end{sideways}}
& Number of utterances         & 1462 &   1786 & 2269 & 588 & 64,254 \\
& Mean (SD) utterance length    & 7.95 (5.46) &   6.79 (3.50) & 6.03 (3.00) & 25.50 (14.63) & 7.16 (4.09) \\
& Number of tokens             & 11,621 &   12,119 & 13,676 & 14,995 & 459,787 \\
& Out-of-vocabulary rate       & 2.21\% &   2.24\% & 3.58\% & 12.04\% & 0.50\% \\
\hline

\end{tabular}
\end{table*}

We explored 5 datasets, three that capture child-directed speech at the level of a single child, one aggregating child-directed speech from multiple children, and one with an equivalent amount of text from the web.

\textbf{SAYCam-S, Sarah and Ellie.} These are three different single-child datasets in our experiments. SAYCam-S is the single child dataset used in \citeA{wang2023finding}. The other two child-directed datasets are two sets of transcribed speech from CHILDES \cite{CHILDES}, each directed to one individual child: Sarah (age ranging from 2;3 to 5;1) from the Brown corpus \cite{brown1973first} and Ellie (age ranging from 0;9 to 5) from the Sakali corpus \cite{beaupoil2015multimodal}. These two datasets, respectively sourced from the North American English and the British English sections of the CHILDES database, capture longitudinal recordings in naturalistic contexts. As shown in Table~\ref{tab:dataset}, these three datasets present similar statistics in terms of vocabulary size, length of utterances and number of tokens. 

\textbf{Wikipedia.} As a comparison, we also have a randomly sampled Wikipedia dataset with a parallel amount of text tokens to Ellie, the child dataset that contains the most tokens. (After filtering sentences with fewer than 2 words, as discussed below in Data Preprocessing, the final token counts varied slightly.) Notably, with its longer average utterance length and more complex content, this Wikipedia set has fewer sentences but a larger vocabulary than the aforementioned child-directed datasets. Detailed statistics can be found in Table~\ref{tab:dataset}.

\textbf{CHILDES.} Finally, as a reference, we incorporated the North American Portion of the CHILDES corpus. It contains aggregated child-directed data with a nearly $6\times$ larger vocabulary and approximately $30\times$ more tokens than the single child datasets. See the detailed statistics in Table~\ref{tab:dataset}.

\subsection{Data Preprocessing}
Built on top of \citeA{yedetore-etal-2023-poor}'s data preprocessing procedure, we excluded children's own utterances to replicate data as similar as possible to the sentences children receive and replaced tokens that appear fewer than 3 times with an \verb|<unk>| token. We split approximately 90\% of each dataset to training, 5\% to validation, and 5\% to testing. We also filter out sentences that contain fewer than 2 words during training and validation. Details of dataset statistics for training and validation can be seen in Table~\ref{tab:dataset}.

\subsection{Model Architectures and Training} 

\citeA{wang2023finding} investigated n-gram models, CBOWs and LSTMs. Our evaluation expands to 6 different model architectures, including GPT-2-style and RoBERTa-style Transformers called BabyBERTa\footnote{
Prior research has shown that a scaled-down version of RoBERTa-base termed BabyBERTa, trained on child-directed data, achieves grammatical knowledge comparable to the full RoBERTa-base on the Zorro benchmark \cite{huebner2021babyberta}. We applied their insights and will refer to our RoBERTa-based Transformer as a BabyBERTa-based Transformer in the following sections.
} \cite{radford2019language, liu2019roberta, huebner2021babyberta}, in addition to LSTMs \cite{hochreiter1997long}. We test two model sizes of each model class. The comprehensive list of model architectures used is detailed in Table~\ref{model_architectures}.



\begin{table}
\begin{center} 
\caption{\textbf{Model Architectures.} $\#$ of trainable parameters are based on the SAYCam-S dataset, with slight variation across datasets due to differences in vocabulary size.}
\label{model_architectures} 
\begin{tabular}{rr}
\hline
              Model & $\#$ of parameters\\
\hline
     LSTM (1-layer) &  3.3M \\
     LSTM (2-layer) &  5.4M \\
    GPT-2 (2-layer) &  7.8M \\
    GPT-2 (8-layer) & 26.7M \\
BabyBERTa (2-layer) &  7.8M \\
BabyBERTa (8-layer) & 26.8M \\
\hline
\end{tabular} 
\end{center} 
\end{table}

\textbf{Training objectives.} All models were trained from scratch. For LSTMs and GPT-2-based Transformers, the models aimed to predict the next token in a short utterance, using cross-entropy loss for training. For the BabyBERTa-based Transformer, the model was trained to predict randomly masked tokens, such that 15\% of the tokens in each utterance were masked anew during each presentation.
\begin{table*}[!ht]
\small
\centering
\caption{\textbf{Validation Perplexity.}}
\label{tab:ppl}
\begin{tabular}{r|rrr|rr}
\hline
Model               &\textbf{SAYCam-S} & \textbf{Sarah}   & \textbf{Ellie}   & Wikipedia & CHILDES \\
\hline
LSTM (1-layer)      &    18.01 &    18.45 &   23.86 &    102.00 &   23.45 \\
LSTM (2-layer)      &    18.47 &    18.40 &   23.59 &    98.70  &   23.74 \\
GPT-2 (2-layer)     &    18.74 &    18.97 &   23.93 &    127.58 &   20.81 \\
GPT-2 (8-layer)     &    18.42 &    18.46 &   23.94 &    130.54 &   20.15 \\
BabyBERTa (2-layer) &    10.41 &    10.96 &   16.24 &     74.38 &   10.39 \\
BabyBERTa (8-layer) &     \textbf{9.25} &    10.67 &   14.94 &     65.10 &   10.35 \\

\hline
\end{tabular}
\end{table*}

\begin{table*}[!ht]
\small
\centering
\caption{\textbf{Zorro Test Accuracies (\%).}}
\label{tab:zorro_acc}
\begin{tabular}{r|rrr|rr}
\hline
Model               & \textbf{SAYCam-S} & \textbf{Sarah}   & \textbf{Ellie}   & Wikipedia & CHILDES \\
\hline
LSTM (1-layer)      &    66.43 &    68.98 &    66.45 &     59.44 &   78.28 \\
LSTM (2-layer)      &    69.18 &    68.25 &    64.59 &     61.64 &   81.49 \\
GPT-2 (2-layer)     &    68.22 &    68.70 &    65.40 &     57.47 &   86.40 \\
GPT-2 (8-layer)     &    65.76 &    \textbf{70.49} &    66.45 &     61.88 &   87.83 \\
BabyBERTa (2-layer) &    69.57 &    70.23 &    66.28 &     59.02 &   84.63 \\
BabyBERTa (8-layer) &    65.45 &    66.42 &    64.46 &     59.54 &   81.65 \\
\hline
\end{tabular}
\end{table*}

\textbf{Model configurations.}
We trained 2 architectures of large and small sizes for each model class, resulting in a total of 6 architectures. These include uni-directional LSTMs (1 layer and 2 layers), as well as GPT-2-based and BabyBERTa-based Transformers (2 layers and 8 layers), as listed in Table~\ref{model_architectures}. Subsequently, we performed an extensive hyperparameter search. We tuned and identified the best hyperparameters based on validation perplexity for each of our five datasets. For the hyperparameter search, we standardized all model embedding and hidden sizes to 512 and all FFN intermediate sizes for Transformer-based models to 2048. We used \verb|ReduceOnPlateau| learning rate scheduler in PyTorch, which reduces the learning rate by a factor of 10 after the validation loss plateaus for 2 consecutive epochs. We used early stopping to select the checkpoint with the best validation loss. 
We tuned other hyper-parameters based on validation performance, including:
\begin{itemize}
  \itemsep0em 
  \item \textbf{learning rate} $ \in \{1 \times 10^{-4},~ 3 \times 10^{-4},~ 1 \times 10^{-3},~ 3 \times 10^{-3}\}$
  \item \textbf{batch size} $ \in \{8, 16, 32\}$
  \item \textbf{weight decay} $ \in \{0.01,~ 0.05, ~0.1, ~ 0.15,~ 0.24\}$
  \item \textbf{dropout rate} $ \in \{0.01, ~0.05, ~0.1, ~0.2, ~0.3, ~0.4, ~0.5\}$
  \item \textbf{number of attention heads} (for Transformer-based models) $ \in \{8, ~16, ~32\}$
  
\end{itemize}
Performance for a particular configuration is averaged across 3 runs with different random seeds. As a measure of generalization, the validation perplexity score\footnote{The perplexity is the exponentiation of the validation cross-entropy loss, defined as:
$\textrm{perplexity} = \exp(\mathrm{H}(X))$, $\mathrm{H}(X)=-\frac{1}{N} \sum_{i=1}^N \log \mathrm{P}\left(x_i\right)$,
where $\mathrm{H}$ is the cross-entropy, and $X$ is a random variable denoting a token. We used it as a more straightforward measure of model performance on next-word prediction tasks.} is shown in Table~\ref{tab:ppl}.

\subsection{Tokenizer}
Simple word-level tokenizers were used to facilitate our analyses of the learned word embeddings (e.g., Fig. \ref{fig:semantic_ategories}), constructed with Hugging Face Tokenizers for each dataset. Refer to Table~\ref{tab:dataset} for the vocabulary size for each dataset.


\section{Results}
We analyze each trained model through linguistic acceptability tests for linguistic knowledge, visualizations of word embeddings for syntactic and semantic category structures, and cloze tests for noun-verb distinction within context. In each analysis, we find robust results similar to \citeA{wang2023finding} across all models with different configurations.

\subsection{Linguistic Acceptability Tests}
Following \citeA{wang2023finding}, we tested models' sensitivity to linguistic knowledge such as subject-verb agreement on the Zorro test suite \cite{huebner2021babyberta}. This test suite evaluates 13 grammatical phenomena on 23 tests, each containing 2000 minimal sentence pairs. To avoid out-of-vocabulary words, \citeA{wang2023finding} filtered out all minimal pairs containing tokens outside of their SAYCam-S vocabulary, left with 15 tests, each containing fewer than 700 pairs. In this work, we regenerated Zorro based on the original linguistic templates and the intersected vocabulary of our 5 datasets, resulting in a full 23 tests.\footnote{The regenerated Zorro test suite can be found in \url{ https://github.com/wwt17/Zorro}.}

\begin{figure*}[!t]
  \centering
  \includegraphics[width=\textwidth]{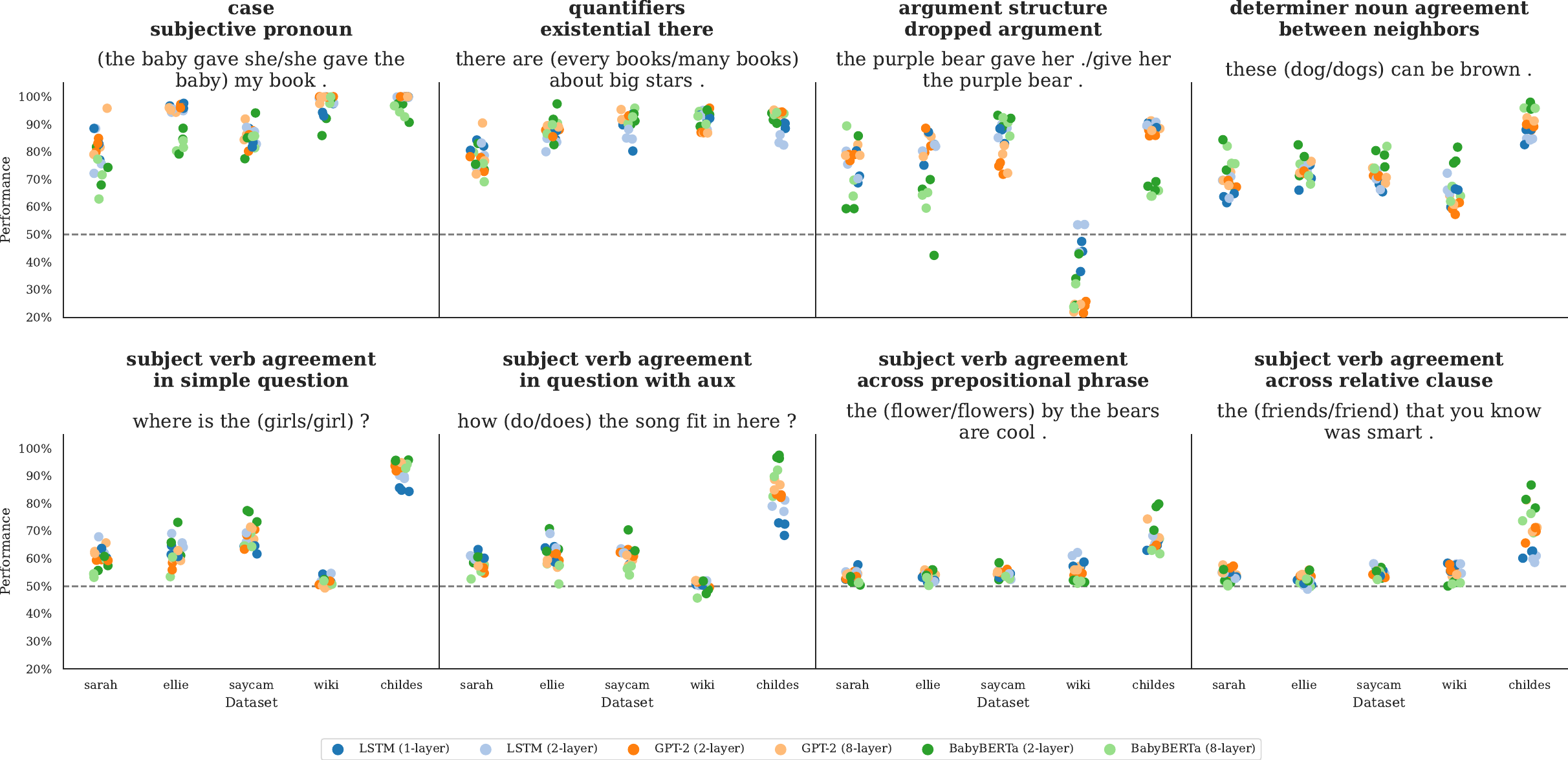}
  \caption{\textbf{Zorro test accuracies across different settings.} We tested 6 model architectures on 23 linguistic tests in Zorro. Each model architecture, trained with 3 seeds, yielded 18 accuracy data points per dataset. Our scatter plots show results for 8 selected tests, with the test name and an example sentence pair (unacceptable/acceptable) highlighted above each. For example, models evaluate which is more acceptable in the ``case--subjective pronoun'' test: ``the baby gave she my book.'' or ``she gave the baby my book.'' We found models trained on single-child datasets excel in specific tests but struggle in others, like subject-verb agreement. Four high-performing tests are shown in the first row, and four lower-performing tests, particularly for subject-verb agreement, are in the second row. Chance is the dotted line. Runs with 3 seeds show variability, similar to previous findings \mycite.}
  \label{fig:Zorro_plot}
\end{figure*}

\textbf{Test accuracy.} From Table~\ref{tab:zorro_acc}, we can see average Zorro test accuracies over 3 different random seeds are consistent among 3 single-child datasets (Sarah, Ellie, and SAYCam-S), nearly all of which reached over 65\% correct (chance is 50\%). Among all single-child-directed datasets, the Sarah dataset trained models with the best Zorro accuracy in all model architectures except the LSTM (2-layer). Comparatively, across all 5 datasets studied, models trained on the Wikipedia dataset exhibit the lowest Zorro accuracy,\footnote{As an example, models trained on the Wikipedia dataset perform the worst on the test for ``argument structure dropped argument'' (as shown in Figure~1, row 1, plot 3), where models are tested on sentences pair such as ``the purple bear gave her./give her the purple bear.'' Since the Wikipedia training dataset does not contain sentences that start with the word ``give'', models yield a high perplexity score on this token and make incorrect judgments.} while those trained on the CHILDES dataset achieve the highest.
Furthermore, for each specific linguistic test, models trained on single child datasets give consistent performances as seen in Figure~\ref{fig:Zorro_plot}. The first row illustrates four linguistic tests where most models trained on single-child datasets perform well, whereas the second row shows models perform poorly on subject-verb agreement.\footnote{A complete plot for model performances on all tests can be found in \url{https://github.com/yuluqinn/single-child-robustness}.}

In particular, all models trained on child-directed datasets exhibit high performance on the ``quantifiers--existential there'' test and perform near chance levels on the ``subject-verb agreement--across relative clause'' test, which aligns to \citeA{wang2023finding} conclusion from previous evaluations. As a comparison, models trained on CHILDES achieve higher test accuracy than models trained on other datasets, yet there is a noticeable variance in their accuracy as shown in the bottom right plot of Figure~\ref{fig:Zorro_plot}. This variability underscores the challenge of mastering the syntactic knowledge required for subject-verb agreement tests, despite the more enriched linguistic context CHILDES provides. More generally, the CHILDES corpus, which is much larger than other datasets, also yielded the best performance in many other tests.

\begin{figure*}[!t]
  \centering
  \includegraphics[width=\textwidth]{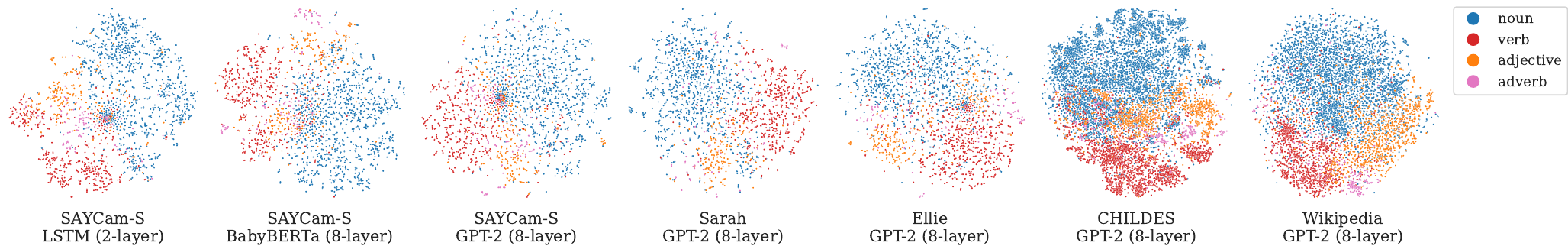}
  \caption{\textbf{Clustering different models' word embeddings for syntactic categories.} We ran t-SNE to visualize embeddings of all words in the vocabulary that are categorized into one of the four syntactic categories: noun, verb, adjective, and adverb. t-SNE uses $1 - \cos(u, v)$ as the distance metric. We show seven visualizations here from various training datasets and model architectures labeled below the plots. Nouns and verbs form two large salient clusters, while adjectives and adverbs are mostly clustered together.}
  \label{fig:syntactic_categories}
\end{figure*}

\subsection{Visualizations for Syntactic and Semantic Categories}

\begin{figure*}[!t]
  \centering
  \begin{subfigure}[b]{0.4\textwidth}
      \centering
      \includegraphics[width=\textwidth]{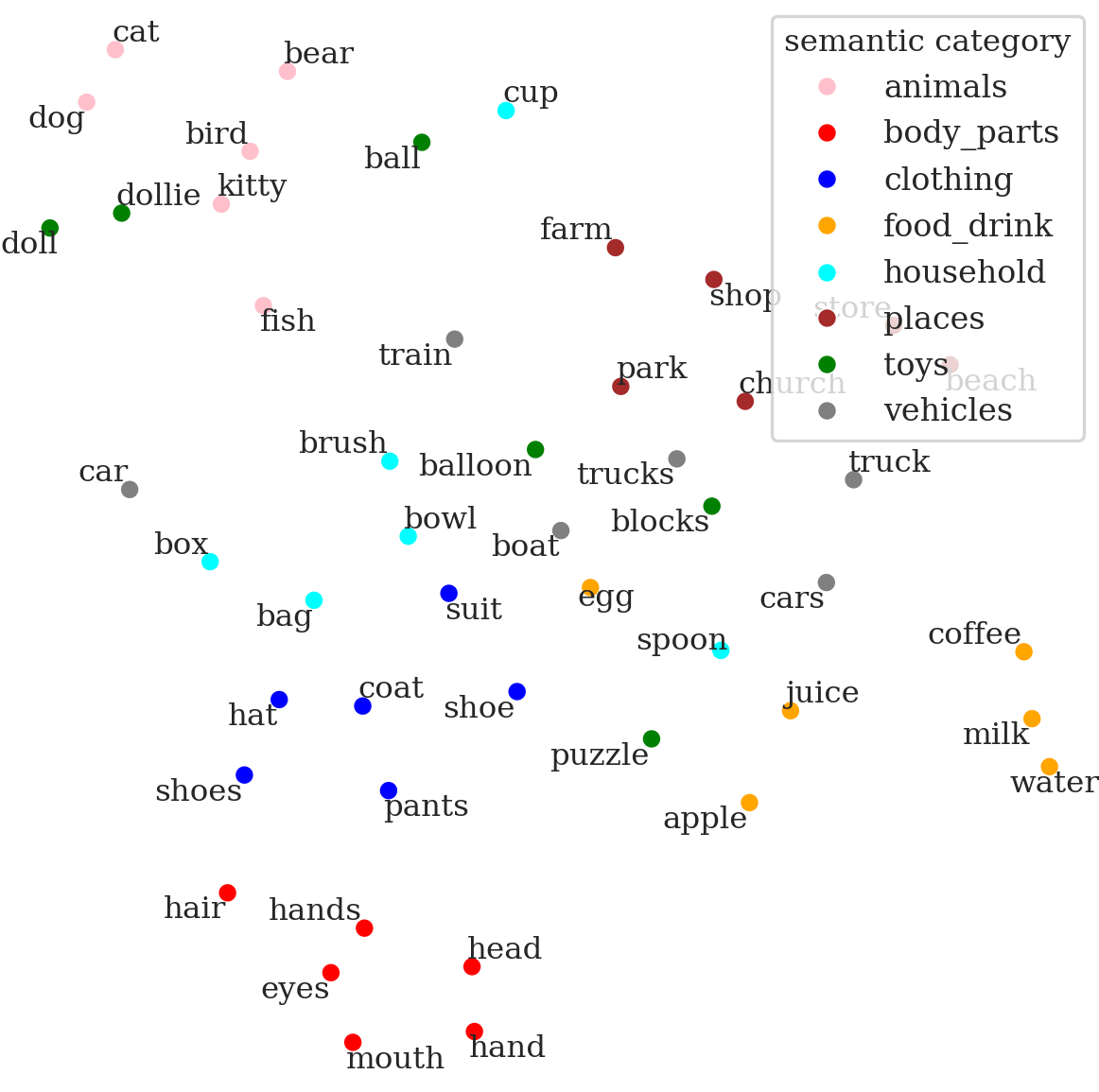}
      \caption{GPT-2 (2-layers) t-SNE}
  \end{subfigure}
  \begin{subfigure}[b]{0.4\textwidth}
      \centering
      \resizebox{.5\textwidth}{!}{
      \includegraphics[width=\textwidth]{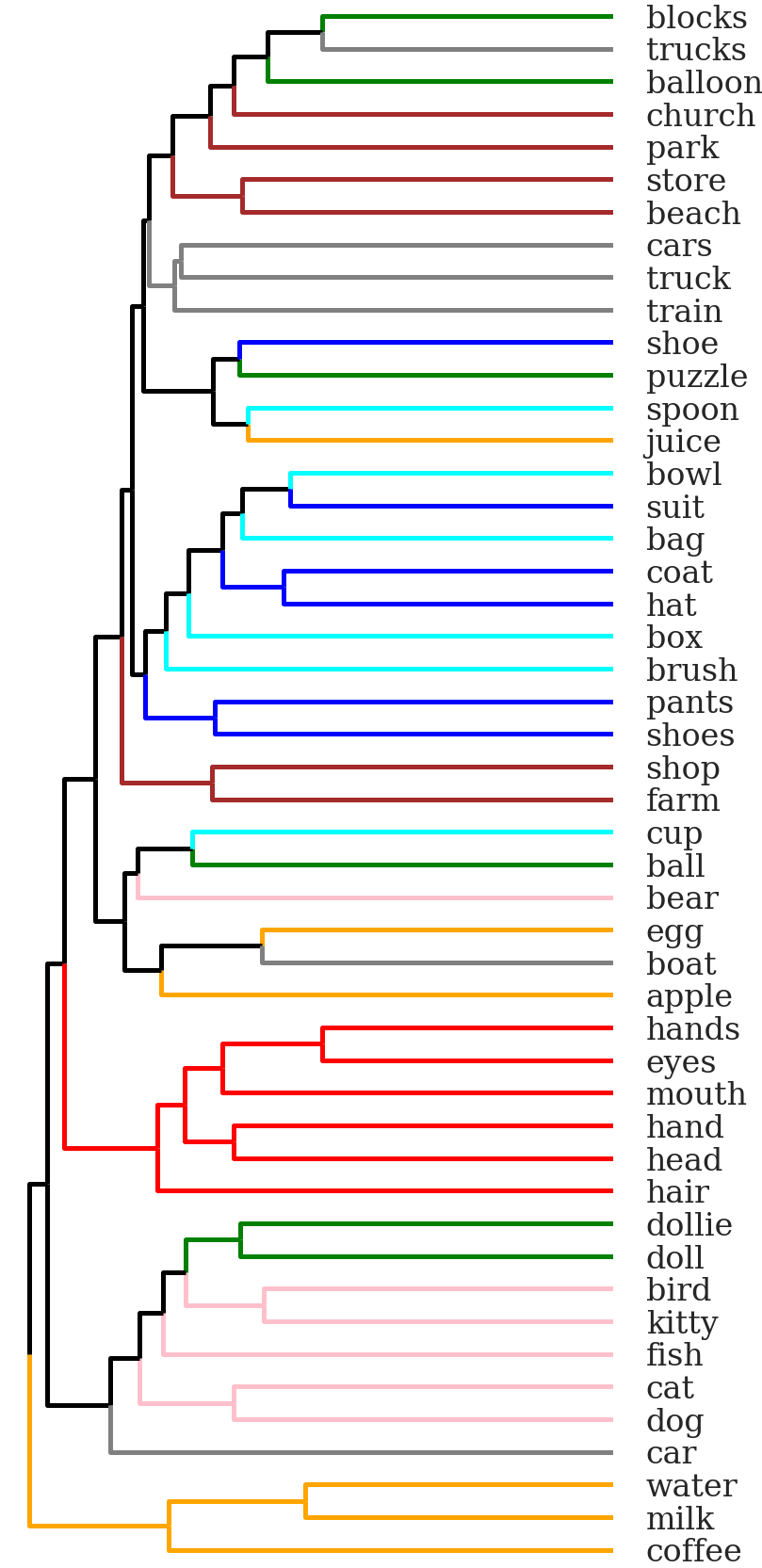}
      }
      \caption{GPT-2 (2-layer) Dendrogram clustering}
  \end{subfigure}
  \\
  \begin{subfigure}[b]{0.35\textwidth}
      \centering
      \includegraphics[width=\textwidth]{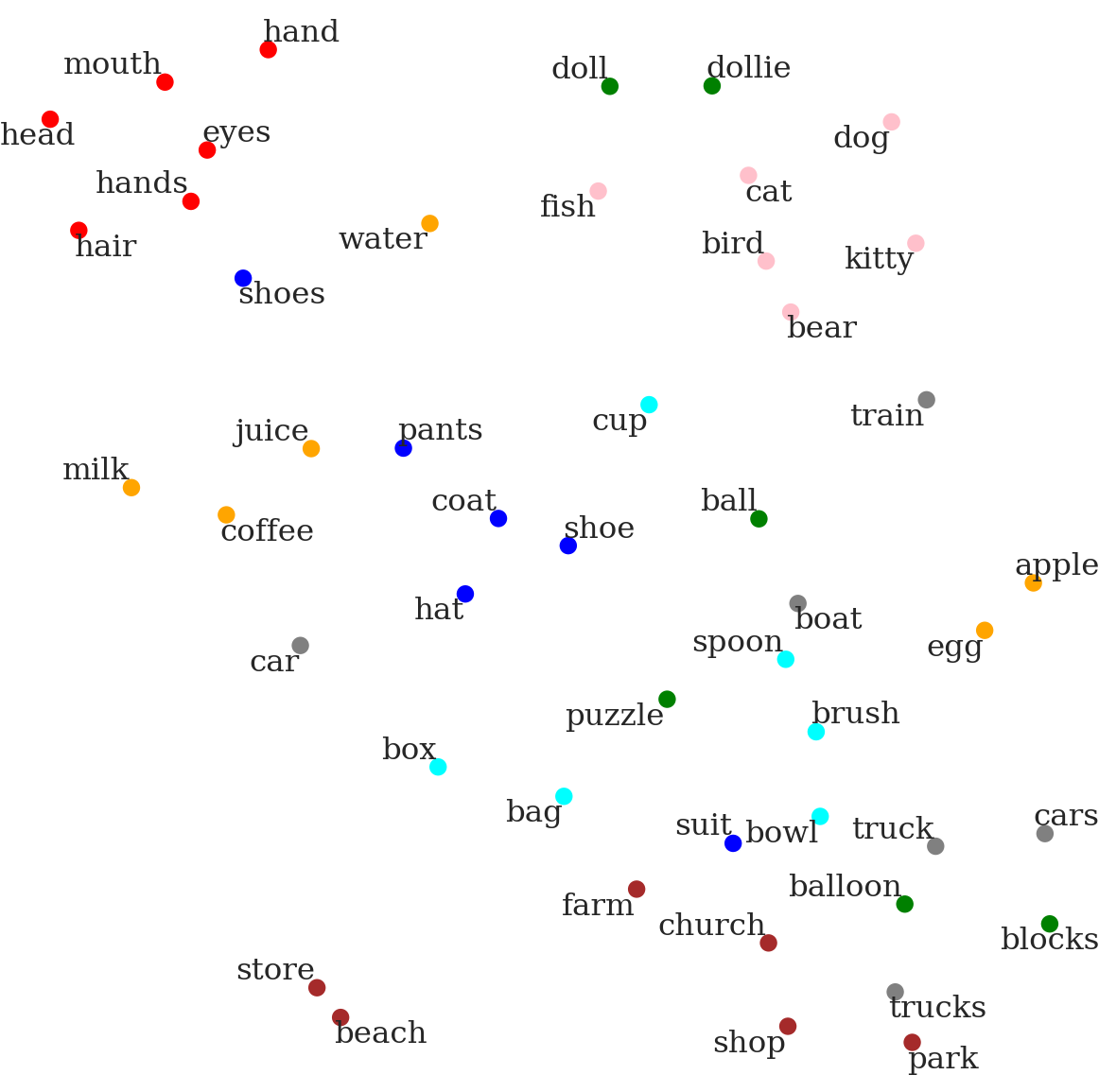}
      \caption{BabyBERTa (2-layer) t-SNE}
  \end{subfigure}
  \hspace{1cm}
  \begin{subfigure}[b]{0.35\textwidth}
      \centering
      \includegraphics[width=\textwidth]{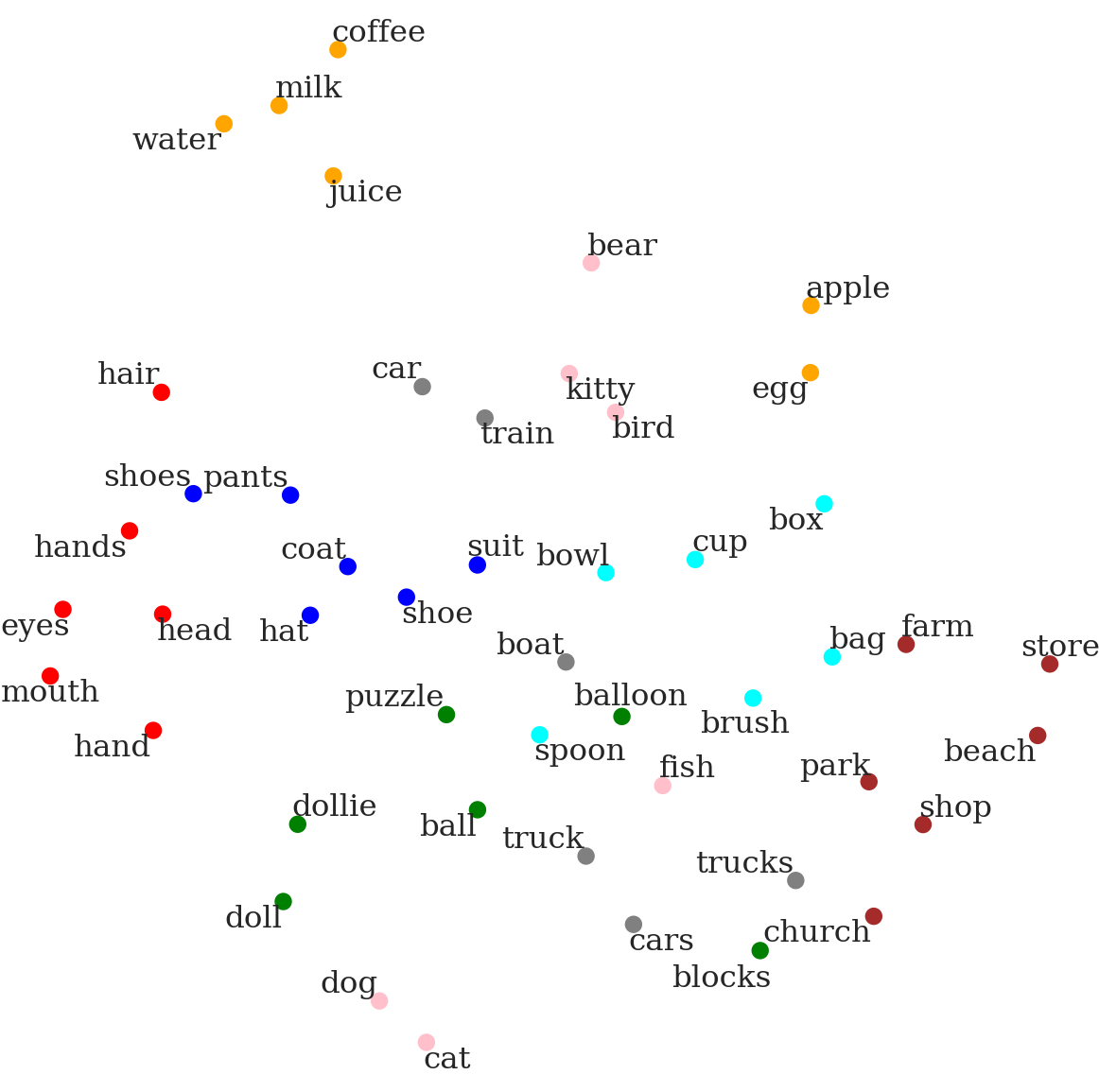}
      \caption{LSTM (1-layer) t-SNE}
  \end{subfigure}
  \caption{\textbf{Clustering word embeddings for semantic categories.} Here we visualize word embeddings of three architectures trained on the Sarah dataset: (a, b) GPT-2 (2-layer), (c) BabyBERTa (2-layer), (d) LSTM (1-layer). Again, t-SNE and dendrogram plots use the cosine measure in Figure~\ref{fig:syntactic_categories}. We present the 6 most frequent words from 8 different categories. There are distinct clusters corresponding to semantic categories, including body parts, clothing, animals, and places.}
  \label{fig:semantic_ategories}
\end{figure*}


In their study, \citeA{wang2023finding} followed a plan of analysis from Elman’s pioneering work \cite{elman1989representation, elman1990finding, elman1991distributed}, demonstrating that CBOW and LSTM models when trained solely on the SAYCam-S dataset can form emergent clusters corresponding to syntactic categories such as nouns, transitive verbs, and intransitive verbs, and semantic categories such as food, animals and body parts. To analyze the cluster structures of word embeddings in their trained models, they visualized the embeddings by t-SNE \cite{van2008visualizing} and cluster dendrograms.

To test the robustness of \citeauthor{wang2023finding}'s findings, our study expands these visualizations to all models we mentioned above.
As for syntactic distinctions, we found all models consistently exhibited clustering patterns in t-SNE plots and dendrograms across various datasets. We first analyze word embeddings of four syntactic categories (nouns, verbs, adjectives, and adverbs) using t-SNE, as illustrated in Figure~\ref{fig:syntactic_categories}. Focusing specifically on the three single-child datasets, we observe a distinct separation between nouns (marked in red) and verbs (marked in blue). Although some overlap exists, clusters of adjectives and adverbs are still discernible. Models trained on CHILDES and Wikipedia datasets displayed more distinct clustering, likely due to their broader vocabularies compared to single-child datasets.

As for semantic categorization, we use the same 8 child-directed semantic categories in \citeA{wang2023finding}, which was derived from WordBank \cite{Frank2016WordbankAO}. Due to differences in vocabulary, we cannot use the same set of words across all datasets. Therefore, for each dataset, we adapt the set of words in each category, enabling visualization of the six most frequent words per category. Figure~\ref{fig:semantic_ategories} displays three models and reveals visually identifiable clusters such as body parts, clothing and animals.

\subsection{Cloze Tests}
\begin{table*}[!ht]
\centering
\caption{\textbf{Cloze test statistics and accuracies (\%) of differentiating noun vs.\ verb.} We build the cloze tests from the validation set for each dataset independently and evaluate the models correspondingly.}
\label{tab:cloze}
\begin{tabular}{r|rrr|rr}
\hline
                    & \textbf{SAYCam-S} & \textbf{Sarah}    & \textbf{Ellie}    & Wikipedia & CHILDES \\
\hline
Number of clozes    &     2412 &     1763 &     1801 &       343 &   74266 \\
Ratio of noun clozes&  35.16\% &  34.66\% &  38.87\% &   69.97\% & 38.76\% \\
\hline
LSTM (1-layer)      &    97.89 &    96.48 &    94.23 &     93.88 &   96.66 \\
GPT-2 (2-layer)     &   \textbf{ 98.09} &    95.92 &    94.39 &     93.88 &   97.23 \\
GPT-2 (8-layer)     &    97.97 &    96.31 &    94.11 &     92.13 &   97.40 \\
BabyBERTa (2-layer) &    96.93 &    95.07 &    93.78 &     93.59 &   97.22 \\
BabyBERTa (8-layer) &    97.51 &    94.55 &    93.73 &     94.75 &   96.33 \\
\hline
\end{tabular}
\end{table*}

In addition to examining emergent lexical classes in the representation space, we wanted to further test if models can properly identify the syntactic category of a missing word based on its surrounding context. Therefore, following \citeA{wang2023finding}, we apply cloze tests \cite{Taylor1953ClozePA} to provide further evidence for syntactic category structures, specifically the noun-verb distinction. We use clozes such as ``we are going to \underline{\smash{\ \ \ \ }} here'', where this cloze expects either a noun or a verb.\footnote{Similar to the category distinction test in \citeA{kim-smolensky-2021-testing}.} We follow the same process as \citeA{wang2023finding} to generate and evaluate the clozes for each dataset. Cloze test statistics and accuracies are shown in Table~\ref{tab:cloze}. All of our models achieve over 90\% accuracy, consistently demonstrating their ability to contextually differentiate nouns and verbs.

\section{General Discussion}
In order to study the robustness of \citeA{wang2023finding}'s learnability results from one child's linguistic input, we systematically trained 6 model architectures on 3 different single-child datasets. We found all trained models achieved consistent results in distinguishing syntactic and semantic categories of words, as well as sensitivity to several linguistic phenomena. We observed high performance on linguistic tests such as quantified existential ``there'' constructions, case of subjective pronouns, and dropped argument for ditransitive verb. But these models consistently failed on more complicated linguistic tests, such as subject-verb agreement across relative clause.

Unlike other work considering the importance of the domain of child-directed speech for learnability, this paper focuses specifically on the role of input to a single child. This approach offers a more realistic baseline than methods that train models on larger, aggregated data sources. With a similar goal, BabyLM challenge \cite{warstadt2023findings} explores learning under limited data conditions. However, even the smallest data track in the BabyLM challenge contains about 40 times more data (10M word tokens) than our single-child dataset. Similarly, in the study by \citeA{huebner2021babyberta}, a RoBERTa-based Transformer was trained on 5M tokens from an age-ordered version of CHILDES \cite{Huebner2020OrderMD} and an equivalent amount from a Wikipedia dataset. Their analysis of the model's performance across various linguistic phenomena was conducted on Zorro. Intriguingly, we observed comparable patterns in our study, even though we used a much smaller dataset comprising single-child linguistic input and a corresponding Wikipedia dataset. Specifically, we found that models trained on the Wikipedia dataset struggled with tests such as dropped argument for ditransitive verb and local attractor in question with auxiliary verb, while the single-child datasets consistently outperformed in these areas. This closely mirrors the findings from \citeA{huebner2021babyberta}'s study using aggregated data sources and larger data quantity. Our results suggest that even limited data can be indicative of differences between datasets and, potentially, that child-directed speech may better equip models with the necessary linguistic abilities for certain tests.

The second key contribution of our study is an in-depth examination of the robustness of the findings by \citeA{wang2023finding}, which were originally based on one single-child dataset: SAYCam-S. We expanded this investigation to include 3 single-child datasets with 2 baselines and 6 model architectures, significantly broadening the scope. Additionally, we enhanced the methodology for linguistic evaluation using the Zorro test suite \cite{huebner2021babyberta}. \citeauthor{wang2023finding} previously limited their analysis to sentence pairs from Zorro that matched SAYCam-S's vocabulary, which resulted in a reduced test scope covering only 15 out of 23 tests and fewer than 700 sentence pairs per test. This limited size potentially weakened the validity of their conclusions.
In contrast, we regenerated the Zorro test suite to align with the intersected vocabulary. Our models were then tested on comprehensive new 23 tests encompassing all 13 linguistic phenomena, with 2,000 sentence pairs in each test. This approach has yielded more robust and reliable results.

Our study demonstrates that models with different configurations can consistently learn to distinguish several syntactic and semantic categories and are sensitive to certain linguistic tests based solely on the linguistic input from a single child. However, we acknowledge several limitations. Firstly, while models demonstrate the ability to form syntactic and semantic clusters distinguishing lexical classes, it remains unclear how they acquire this representation and whether their understanding of these categories aligns with human cognition. Secondly, our evaluation methods, though insightful, are not exhaustive. The behavioral tests using Zorro are valuable for assessing responses to grammatical variations in sentences. However, it is important to note that Zorro has its limitations \cite{genben}, and we still lack more systematic semantic evaluations. Lastly, our models are exclusively trained on transcribed speech. \citeA{wang2023finding} and \citeA{warstadt2023findings} suggest that integrating multiple modalities given realistic experience is a significant challenge in language learning, although there has been recent progress \cite{vong2024grounded}. We see multi-modal learning as a promising means of enhancing model data efficiency and realism by better capturing the learning problem faced by a young child.

\section{Acknowledgments}
We thank Wai Keen Vong, Cara Leong, Cindy Luo and Solim LeGris for helpful feedback on earlier drafts.  This work was supported by NSF Award 1922658 NRT-HDR: FUTURE Foundations, Translation, and Responsibility for Data Science.




\nocite{ChalnickBillman1988a}
\nocite{Feigenbaum1963a}
\nocite{Hill1983a}
\nocite{OhlssonLangley1985a}
\nocite{Matlock2001}
\nocite{NewellSimon1972a}
\nocite{ShragerLangley1990a}

\bibliographystyle{apacite}

\setlength{\bibleftmargin}{.125in}
\setlength{\bibindent}{-\bibleftmargin}

\bibliography{CogSci,library_clean}

\end{document}